\documentclass[conference]{IEEEtran}
\IEEEoverridecommandlockouts
\usepackage{cite}
\usepackage{amsmath,amssymb,amsfonts}
\usepackage [algoruled,linesnumbered]{algorithm2e}
\usepackage{graphicx}
\usepackage{textcomp}
\usepackage{xcolor}
\usepackage{todonotes}
\usepackage{caption}
\usepackage{subcaption}
\usepackage{multicol}
\usepackage{multirow}
\usepackage[symbol]{footmisc}

\usepackage[square, numbers, comma, sort&compress]{natbib}

\def\BibTeX{{\rm B\kern-.05em{\sc i\kern-.025em b}\kern-.08em
    T\kern-.1667em\lower.7ex\hbox{E}\kern-.125emX}}
\begin{document}

\title{Contextual Bandits Evolving Over Finite Time\\
}

\author{\IEEEauthorblockN{Harsh Deshpande$^*$}
\IEEEauthorblockA{\textit{Department of Electrical Engineering} \\
\textit{Indian Institute of Technology}\\
Mumbai, Maharashtra \\
hdeshpande5998@gmail.com}
\and
\IEEEauthorblockN{Vishal Jain$^*$}
\IEEEauthorblockA{\textit{Department of Electrical Engineering} \\
\textit{Indian Institute of Technology}\\
Mumbai, Maharashtra \\
vishalj409@gmail.com}
\and
\IEEEauthorblockN{Sharayu Moharir}
\IEEEauthorblockA{\textit{Department of Electrical Engineering} \\
\textit{Indian Institute of Technology}\\
Mumbai, Maharashtra \\
sharayum@ee.iitb.ac.in}
}

\maketitle

\begin{abstract}
Contextual bandits have the same exploration-exploitation trade-off as standard multi-armed bandits. On adding positive externalities that decay with time, this problem becomes much more difficult as wrong decisions at the start are hard to recover from. We explore existing policies in this setting and highlight their biases towards the inherent reward matrix. We propose a rejection based policy that achieves a low regret irrespective of the structure of the reward probability matrix.
\end{abstract}


\footnotetext{$^*$Joint First Authors}

\section{Introduction}
In the context of restaurant recommendation systems, users can generally be classified into multiple user types with different preferences for different restaurants. Another behaviour that can be observed is that based on reviews provided by past users, the proportion of users preferring one restaurant over the other can change with time.\\
We consider such a setting in which the users/customers are classified into a number of customer types based on which of the restaurants they like the most. We propose an algorithm for a recommendation platform such that whenever a new user comes to the platform, the platform suggests one of the restaurants to the user, and a binary reward is generated based on the reviews provided by the user. The platform is aware of the type of the incoming user but is unaware of the user-restaurant reward probabilities. Further, if a positive reward is generated on being recommended a particular restaurant, the population of people preferring that particular restaurant increases. This can lead to a self-reinforcing behavior that is termed as positive externalities \cite{katz}. This increase in population is modelled as decaying with time, which is intuitive, as over time, the effects of recommendations generally saturate, leading to an equilibrium in the population distribution of customers.\\
We model this setting as an evolving contextual bandits problem where the user type is regarded as the context and the restaurants are modelled as the bandit arms. The population distribution of the context changes according to the arms pulled and the corresponding rewards accrued. A trivial way to maximise the total reward accrued in such a setting would be to keep showing the arm with the maximum probability of being accepted irrespective of the context. Although such a policy guarantees minimum regret over the infinite time horizon, it does not guarantee minimum regret over a short time period which is usually the case in such settings. Moreover, because of the decaying nature of the externalities, suboptimal decisions in the beginning can lead to an increase in regret which might be difficult to compensate under a short time horizon.


\section{Previous Work}
Contextual bandits have been explored in various works \cite{auernon, langford, li2010}. Much work has been done on a non-evolving setting where the incoming population of the different contexts is not affected by the arms pulled or the rewards accrued. \\
Evolving bandits have been explored by  \cite{Virag} who have developed policies to minimize regret in a similar setting. However, they highlight that their model is different from contextual bandits. Furthermore, in contrast to their setting of externalities, we have an evolution that decays with time, thus making the problem more difficult as wrong decisions at the start are harder to correct.\\

\section{Setting}
\subsection{Context and Arm Rewards}
Let $\{1,2,....n\}$ be the types of context that can arrive at any time instant. Let the set of arms be $\{1,2,3....m\}$ (m $\geq$ n). At each time instant, the context is sampled from a distribution $d(t)$. Here, $d(t)$ is a $n$x1 array where $d_i(t)$ denotes the population density of customer type $i$ at time $t$. For each such context, an arm is pulled and the obtained reward is 0 or 1. \\
Let reward obtained at time $t$ be $r_{t}$. The cumulative reward till time $t$ is defined as $R_{t} = \sum_{k=1}^{t}r_{k}$. Similarly $R_{i,j,t}$ denotes reward accrued till time $t$ by pulling arm $j$ on arrivals of type $i$. Also, $T_{i,j,t}$ and $S_{i,j,t}$ denote the number of times arm $j$ was pulled on arrival of context $i$ and the number of such instances with 0 reward (ie, the number of times the arm was rejected for that user type), respectively. Thus, $T_{i,j,t} = R_{i,j,t}+S_{i,j,t}$ for all $i$ and $j$.

\subsection{Reward Probabilities}
The probability of getting reward 1 for context $i$ and arm $j$ is equal to $u_{ij}$. Thus, the reward probabilities can be compactly represented by the matrix:\\

$ M = 
\begin{bmatrix}
\mu_{11} & \mu_{12} & .. & \mu_{1m} \\
\mu_{11} & \mu_{12} & .. & \mu_{2m} \\
.. \\
\mu_{n1} & \mu_{n2} & .. & \mu_{nm}
\end{bmatrix}$ \\

In $M$, the maximum element of each row is the diagonal entry corresponding to that row - we call this the "maxima along the diagonal" structure. This allows every user type to have a unique "most-preferred" arm. Thus, $u_{ii} \geq u_{ij}  \forall {i,j}$. Further, without loss of generality, we arrange the rows in decreasing order of their highest elements. Thus, $u_{ii} \geq u_{jj} \forall i \leq j$.

\subsection{Evolution of $d(t)$}
We consider a setting where the population distribution of user types changes only when the reward accrued is 1. Thus, at time instant $t$ if context $i$ arrived, arm $j$ was pulled and reward was $r$, $d(t)$ is updated as:
\begin{align*}
   d_j(t+1) &= d_j(t) + \frac{\delta*r}{\sqrt{t}} \\
   d(t+1) &= \frac{d(t+1)}{\sum_{k=1}^{n}d_k(t+1)} \tag*{(Normalization)}
\end{align*}
where $\delta$ is a constant indicative of the step-size.\\
We can see that any non-decreasing function of $t$ can be used instead of $\sqrt{t}$. We restricted ourselves to functions of the form $t^\frac{1}{b}$ as they form ODEs that can be solved in closed form. Further we chose $b$ to be 2 as it was high enough to have appreciable change in the distribution and low enough to guarantee alpha will saturate.

\section{Policies}
We have explored various policies for different settings. In this section, we describe in short the policies and then introduce our own policy Reward Based Arm Elimination (RBAE) at the end.

\subsection{Oracle}
It is easy to see that if we know the underlying reward matrix, in an infinite time horizon, the best policy would be to pull the arm with highest reward probability (for all arms for all contexts). Thus for any context, arm $j$ is pulled where $j$ is $argmax_{j}(max_i(\mu_{ij}))$. 

\subsection{Greedy-Oracle}
This policy assumes knowledge of the "maxima along the diagonal" structure of the probability matrix $M$, and always recommends the best arm for user type $i$, which is $argmax_{j}(\mu_{ij})$.

\subsection{Random Explore then Commit (REC)}
For $t \leq \tau$ (pre-defined), sample arms uniformly at random (exploration). After $\tau$, sample $argmax_{j}R_{i_t,j,\tau}$, i.e., the arm which accrues the highest reward for that user type (exploitation).

\subsection{Balanced Exploration (BE)}
\cite{Virag} proposes an algorithm that structures exploration (in contrast with REC) by balancing exploration across arms, by ensuring every arm accrues at least a minimum reward before deciding on the optimal arm. We implement a version of this modified to our setting where exploration is done across context types.\\ 
\begin{algorithm}
\SetAlgoLined
 $minR = \alpha ln(T)$ \\
 $t = 1$ \\
 \While{$t \leq T$}{
  $i_t = customer\_type(t) $\\
  \eIf{$\exists j \; R_{i_t,j,t} < minR $}{
  Sample arm $a = argmin_jR_{i_t,j,t}$  \\
  $\tau = t$\\
  }{
  Sample arm $a^* = argmin_jT_{i_t,j,\tau}$ \\
  }
  $t=t+1$
  }
\caption{Balanced Exploration}
\label{alg:BE}
\end{algorithm}

\subsection{Rejection-Based Arm Elimination (R-BAE)}
A problem with BE is that when the reward probabilities are low, the exploration phase would take a longer time to complete, thus possibly increasing the regret especially when the time horizon of interest is small. To overcome this limitation, we propose a rejection-based policy where sub-optimal arms for a user type are eliminated when the number of rejections for an arm cross a threshold. We believe that by using such a policy, highly sub-optimal arms would be discarded at the earliest, thereby decreasing the accumulated regret.\\ 
\begin{algorithm}
\SetAlgoLined
 $maxS = \beta ln(T)$ \\
 $A_{i} = \{j|0<j \leq m\} \forall i$\\
 $t = 1$ \\
 \While{$\exists i \; |A_{i}(t)|\neq 1 $}{
  $i_t = customer\_type(t) $\\
  \eIf{$|A_{i_t}|>1$}{
   Sample any arm a $\in |A_{i_t}|$  (uniformly at random)\\
   $A_{i_t} = \{j | S_{i_t,j,t} \leq maxS\}$ \\
   }{
   Sample arm a $\in A_{i_t}$ \\
  }
  $t=t+1$ \\
 }
 \While{$t \leq T$}{
 $i_t = customer\_type(t)$ \\
 Sample arm a $\in A_{i_t}$ \\
 $t = t+1$
 }
 \caption{Rejection-Based Arm Elimination}
 \label{alg:RBAE}
\end{algorithm}





\begin{figure*}
\centering
\begin{subfigure}{.33\textwidth}
  \centering
  \includegraphics[width=\linewidth]{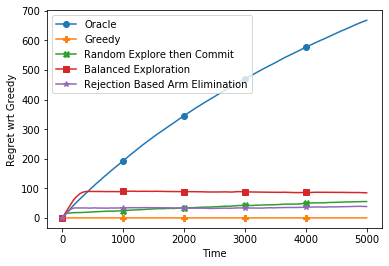}
  \caption{$M$ = [[0.8,0.4],[0.2,0.7]]; $d(1)$ = [0.5,0.5]}
  \label{fig1:sub1}
\end{subfigure}%
\begin{subfigure}{.33\textwidth}
  \centering
  \includegraphics[width=\linewidth]{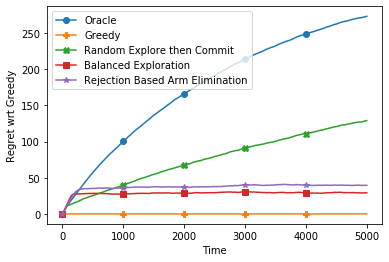}
  \caption{$M$ = [[0.9,0.6],[0.5,0.8]]; $d(1)$ = [0.5,0.5]}
  \label{fig1:sub2}
\end{subfigure}%
\begin{subfigure}{.33\textwidth}
  \centering
  \includegraphics[width=\linewidth]{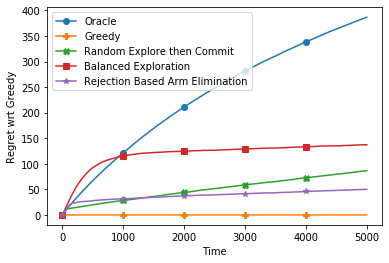}
  \caption{Random $M$; $d(1)$ = [0.5,0.5]}
  \label{fig1:sub3}
\end{subfigure}%
\caption{Aggregate Regret for a fixed $d(1)$}
\label{fig:regret_figure}

\begin{subfigure}{.33\textwidth}
  \centering
  \includegraphics[width=\linewidth]{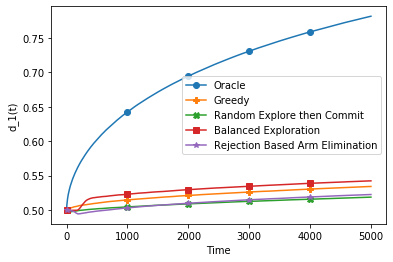}
  \caption{$M$ = [[0.8,0.4],[0.2,0.7]]; $d(1)$ = [0.5,0.5]}
  \label{fig2:sub1}
\end{subfigure}%
\begin{subfigure}{.33\textwidth}
  \centering
  \includegraphics[width=\linewidth]{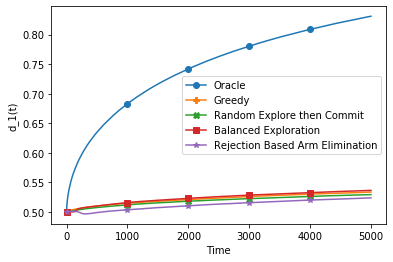}
  \caption{$M$ = [[0.9,0.6],[0.5,0.8]]; $d(1)$ = [0.5,0.5]}
  \label{fig2:sub2}
\end{subfigure}%
\begin{subfigure}{.33\textwidth}
  \centering
  \includegraphics[width=\linewidth]{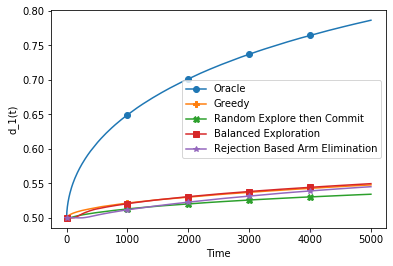}
  \caption{Random $M$; $d(1)$ = [0.5,0.5]}
  \label{fig2:sub3}
\end{subfigure}%
\caption{Evolution of $d_1(t)$ for a fixed $d(1)$}
\label{fig:alpha_figure}
\end{figure*}


\section{Simulations and Discussion}
\begin{figure*}
\begin{subfigure}{.33\textwidth}
  \centering
  \includegraphics[width=\linewidth]{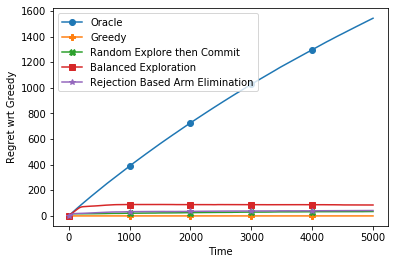}
  \caption{$M$ = [[0.8,0.4],[0.2,0.7]]; $d(1)$ = [0.1,0.9]}
  \label{fig3:sub1}
\end{subfigure}%
\begin{subfigure}{.33\textwidth}
  \centering
  \includegraphics[width=\linewidth]{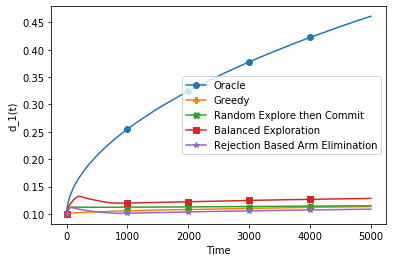}
  \caption{$M$ = [[0.8,0.4],[0.2,0.7]]; $d(1)$ = [0.1,0.9]}
  \label{fig3:sub2}
\end{subfigure}%
\begin{subfigure}{.33\textwidth}
  \centering
  \includegraphics[width=\linewidth]{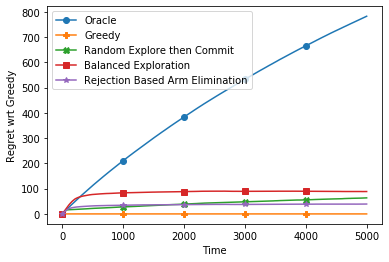}
  \caption{$M$ = [[0.8,0.4],[0.2,0.7]]; Random $d(1)$}
  \label{fig3:sub3}
\end{subfigure}%
\caption{Regret and Evolution of $d_1(t)$ for a fixed $M$}
\label{fig:dist_figure}

\end{figure*}

For simulations, we choose 2 types of context and 2 bandit-arms i.e. $m=n=2$. 500 iterations were run with each iteration lasting for $T=5000$ time instances. The step size of the distribution update $\delta$ was chosen as 0.01. For REC, $\tau = \sqrt{T}$ was used as the exploration time. For BE and RBAE, the thresholds $minR$ and $maxS$ were both taken as $3ln(T)$.\\
Figures \ref{fig:regret_figure} and \ref{fig:alpha_figure} show the aggregate regret and evolution of $d_1(t)$ over time, respectively, with the initial distribution fixed as $d(1)=[0.5,0.5]$, for different values of $M$.\\
\ref{fig1:sub1} uses a probability matrix with sufficient difference between the reward probabilities of the optimal and sub-optimal arms. This leads to BE accumulating a large regret in the exploration phase as it keeps sampling sub-optimal arms till they reach a minimum desired reward. On the other hand, REC is second best and RBAE performs the best in terms of accrued regret. \ref{fig1:sub2} uses a probability matrix with relatively high probabilities of reward for all arms for all contexts. We see that in this case, BE and RBAE perform much better than REC. This can be attributed to the small difference in rewards of optimal and sub-optimal arm thus leading to REC making wrong decisions more often. \ref{fig1:sub3} shows average accumulated regret of the policies across 1250 iterations with the probability matrix randomly changed after every 50 iterations. This was done to remove the biases that the policies had towards certain types (relative values) of the probability matrix. In this case, RBAE performs the best closely followed by REC and then by BE. Note that the "Oracle" always achieves a higher regret in small time horizons as it trades off regret to increase the distribution of the context with the highest possible expected reward. This can be seen in the plots of Figure \ref{fig:alpha_figure}, where Oracle increases $d_1$ to a significantly high value as compared to all the other policies.\\
Figures \ref{fig3:sub1} and \ref{fig3:sub2} show the regret and evolution of $d_1(t)$ for a different value of initial distribution $d(1)$, this time starting with a low value of $d_1 = 0.1$. This can correspond to a setting where a new restaurant enters the market, with a low proportion of customers preferring the entrant initially. \ref{fig3:sub3} shows aggregate regret for the same $M$, but averaged over random values of $d(1)$. Again, RBAE outperforms BE and REC.

\section{Conclusions and Future Work}
We present a new policy "Rejection-Based Arm Elimination" and demonstrate its efficacy in a decaying positive externality setting as compared to previously known policies. This policy eliminates arms based on individual rejections accrued, thereby performing better in terms of acquired regret irrespective of the inherent reward probabilities. We also demonstrate that the other policies can perform well when the probability matrix satisfies certain conditions whereas R-BAE performs well in all cases.\\
In future work, we plan to examine and exploit the correlation between the optimal arms of different customer types and use this extra information to improve expected reward.

\bibliography{COMSNETS}

\begin{thebibliography}{1}
\providecommand{\url}[1]{#1}
\csname url@samestyle\endcsname
\providecommand{\newblock}{\relax}
\providecommand{\bibinfo}[2]{#2}
\providecommand{\BIBentrySTDinterwordspacing}{\spaceskip=0pt\relax}
\providecommand{\BIBentryALTinterwordstretchfactor}{4}
\providecommand{\BIBentryALTinterwordspacing}{\spaceskip=\fontdimen2\font plus
\BIBentryALTinterwordstretchfactor\fontdimen3\font minus
  \fontdimen4\font\relax}
\providecommand{\BIBforeignlanguage}[2]{{%
\expandafter\ifx\csname l@#1\endcsname\relax
\typeout{** WARNING: IEEEtran.bst: No hyphenation pattern has been}%
\typeout{** loaded for the language `#1'. Using the pattern for}%
\typeout{** the default language instead.}%
\else
\language=\csname l@#1\endcsname
\fi
#2}}
\providecommand{\BIBdecl}{\relax}
\BIBdecl

\bibitem{katz}
M.~L. Katz and C.~Shapiro, ``Systems competition and network effects,''
  \emph{Journal of economic perspectives}, vol.~8, no.~2, pp. 93--115, 1994.

\bibitem{auernon}
P.~Auer, N.~Cesa-Bianchi, Y.~Freund, and R.~E. Schapire, ``The nonstochastic
  multiarmed bandit problem,'' \emph{SIAM journal on computing}, vol.~32,
  no.~1, pp. 48--77, 2002.

\bibitem{langford}
J.~Langford and T.~Zhang, ``The epoch-greedy algorithm for contextual
  multi-armed bandits,'' in \emph{Proceedings of the 20th International
  Conference on Neural Information Processing Systems}.\hskip 1em plus 0.5em
  minus 0.4em\relax Citeseer, 2007, pp. 817--824.

\bibitem{li2010}
L.~Li, W.~Chu, J.~Langford, and R.~E. Schapire, ``A contextual-bandit approach
  to personalized news article recommendation,'' in \emph{Proceedings of the
  19th international conference on World wide web}.\hskip 1em plus 0.5em minus
  0.4em\relax ACM, 2010, pp. 661--670.

\bibitem{Virag}
V.~Shah, J.~H. Blanchet, and R.~Johari, ``Bandit learning with positive
  externalities,'' in \emph{NeurIPS}, 2018.

\end{thebibliography}
\bibliographystyle{IEEEtran}

\end{document}